\documentclass[letterpaper, 10 pt, conference]{ieeeconf}  

\IEEEoverridecommandlockouts                              

\usepackage{amsmath} 
\usepackage{amssymb}  
\usepackage{tikz}
\usetikzlibrary{positioning, arrows.meta, quotes}
\usetikzlibrary{shapes,snakes}
\usetikzlibrary{bayesnet}
\tikzset{>=latex}
\tikzstyle{plate caption} = [caption, node distance=0, inner sep=0pt,
below left=5pt and 0pt of #1.south]
\usepackage{algorithmic}
\usepackage{graphicx}
\usepackage{subcaption}
\usepackage{changepage}
\usepackage{textcomp}
\usepackage{xcolor}
\usepackage{url}
\usepackage{multirow}
\usepackage[bookmarks=true]{hyperref} 
\usepackage{array}
\usepackage{booktabs}
\usepackage{tabularray}
\usepackage{makecell}
\usepackage{times}
\usepackage{comment}
\usepackage{cite}
\usepackage{balance}

\title{Remote Surfaces at Your Fingertips: Electrovibration-Based\\Tactile Feedback for Robot Teleoperation via Touchscreen Interfaces}

\author{Alperen Kenan$^{1}$, Juan José García Cárdenas$^{2}$ , Adriana Tapus$^{2}$, Paul Bremner$^{1}$ and Manuel Giuliani$^{3}$
\thanks{$^{\dagger}$This work was supported by the European Commission's Marie Skłodowska-Curie Actions (MSCA) Project RAICAM (GA 101072634), and UK Research and Innovation (UKRI) grant number EP/X025977/1.}
\thanks{For the purpose of open access, the author has applied a Creative Commons Attribution (CC BY) licence to any Author Accepted Manuscript version arising.}
\thanks{$^{1}$Alperen Kenan is a PhD candidate and Paul Bremner is an Associate Professor in Human-Robot Interaction at the Bristol Robotics Laboratory, University of the West of England, Bristol, United Kingdom {\tt\small alperen.kenan@uwe.ac.uk, paul2.bremner@uwe.ac.uk}}
\thanks{$^{2}$ Juan José García Cárdenas is a PhD candidate and Adriana Tapus is Full Professor in the Computer Science and System Engineering Department (U2IS), Autonomous Systems and Robotics Lab, ENSTA, Institut Polytechnique de Paris, Paris, France {\tt\small juan-jose.garcia@ensta.fr; adriana.tapus@ensta.fr}}
\thanks{$^{3}$Manuel Giuliani is a Professor at Kempten University of Applied Sciences, Kempten, Germany {\tt\small manuel.giuliani@hs-kempten.de}}}

\begin{document}

\maketitle
\thispagestyle{empty}
\pagestyle{empty}

\begin{abstract}

Enabling operators to perceive and interact with remote environments naturally is a fundamental challenge in robotic teleoperation. This is especially critical in tasks involving physical interaction, where real-time haptic awareness improves operational safety and effectiveness. Existing kinesthetic haptic feedback methods suffer from instability during rigid surface contacts and remain sensitive to communication delays, while visual cue-based force feedback imposes additional cognitive load and limits sustained situational awareness. This work presents a teleoperation interface that conveys remote surface interactions to the operator through electrovibration-based tactile feedback, enabling naturally mapped force reflection while avoiding the stability issues associated with kinesthetic feedback and the latency limitations of mechanical actuators.

A user study ($N=21$) evaluated interface usability, sense of presence, and operator workload under two force reflection conditions: \textit{visual feedback} and \textit{electrovibration-based tactile feedback}. Characterisation experiments further assessed path-following accuracy and response time across both conditions. Results show that tactile feedback significantly reduced response time by 15.35\% ($p=0.002$, $d=0.96$) and increased the sense of presence by 31\% ($p<0.001$, $d=0.90$) compared to visual feedback, while imposing comparable workload and usability across both conditions.

These findings demonstrate that electrovibration-based tactile feedback is a viable and effective modality for robot teleoperation, improving operator responsiveness and sense of presence in contact-rich manipulation tasks, with direct applicability to safety-critical domains such as nuclear maintenance.

\end{abstract}

\section{Introduction}

Robotic teleoperation enables human operators to perform tasks in environments that are hazardous, remote, or otherwise inaccessible, with applications ranging from nuclear decommissioning and space exploration to surgical robotics \cite{sheridan1992telerobotics, hokayem2006bilateral}. Among these, surface interaction tasks such as swab sampling in nuclear facilities demand precise force control, accurate path-following, and sustained operator awareness of robot-environment contact \cite{johnson2022swab, johnson2021force}. Current interfaces rely heavily on operator skill and impose cognitive demands, particularly when precise force control is required \cite{kenan2025robot}.

Bilateral teleoperation, in which feedback from the remote environment is provided to the operator, offers a pathway towards achieving telepresence \cite{hokayem2006bilateral}, defined as the sensation of being physically present in the remote environment \cite{draper1998telepresence}. Kinesthetic feedback is the most established approach, but suffers from instability during stiff environment contacts and sensitivity to communication delays \cite{lawrence1992stability}. Visual cue-based force feedback avoids these issues but requires interpreting an additional sensory channel, increasing cognitive load and reducing situational awareness \cite{koritnik2010comparison}. Tactile feedback offers an alternative by conveying interaction forces directly to the operator's fingertip in the same modality as the physical interaction, avoiding both the stability limitations of kinesthetic feedback and the cognitive overhead of sensory substitution \cite{pacchierotti2017wearable}. As both visual and tactile feedback remain stable under rigid contact, this study compares these two modalities rather than benchmarking against kinesthetic devices.

Among tactile feedback methods, electrovibration is well suited to touchscreen-based interfaces. It operates by modulating the friction force at the fingertip through an alternating electric voltage on a conductive screen surface \cite{bau2010teslatouch, vardar2017roughness}, with low signal latency ensuring transparent force reflection. Furthermore, electrovibration can render a range of distinguishable tactile sensations \cite{sadia2022exploration}, and its potential integration with tactile robotic perception sensors \cite{lepora2026tactile} on the end effector offers a promising route toward full tactile telepresence. However, its use for reflecting remote surface interactions during teleoperation remains unexplored.

\begin{figure}[t]
\centering
\includegraphics[width=1.00 \columnwidth]{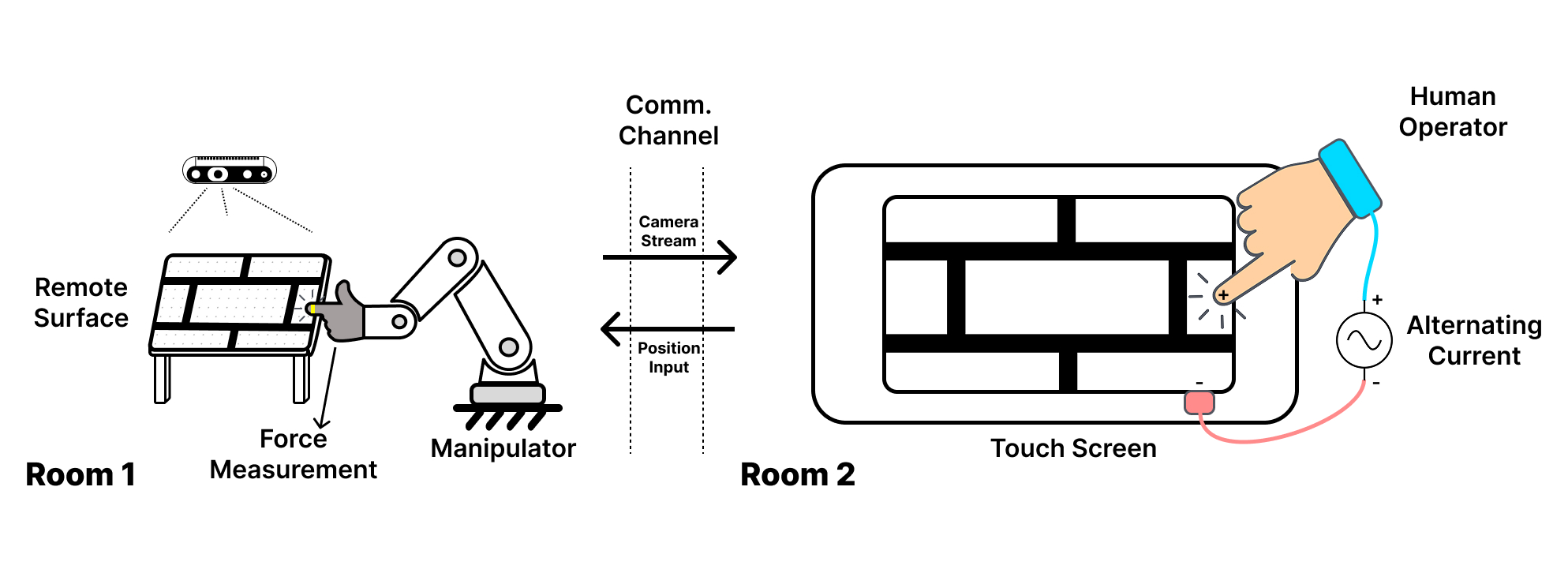}
\caption{Illustration of the touchscreen teleoperation interface}
\label{fig:illustration}
\vspace{-15pt}
\end{figure}

To address the limitations of existing approaches, this paper presents a novel touchscreen-based teleoperation interface that maps remote robot-environment interaction forces to electrovibration-based tactile feedback with negligible delay. The interface aims to: (i) reduce operator workload through intuitive control mappings that emulate direct interaction with the environment; (ii) improve safety and task performance by enabling real-time awareness of robot-environment interaction during surface manipulation; and (iii) enhance the sense of telepresence, so that operators feel as though they are directly interacting with the remote environment from the safety of a control room.

Figure \ref{fig:illustration} illustrates the working principle of the teleoperation setup. The planar position of the operator's finger on the touchscreen is mapped to the robot end-effector coordinates, while a live camera stream supports intuitive control. The interface conveys sensor readings to the operator via electrovibration-based tactile feedback, reflecting remote interaction forces directly to the fingertip.

This paper makes three primary contributions: (1) the implementation of electrovibration as a tactile feedback method for reflecting robot-environment interaction forces during teleoperation without notable delays; (2) the design, characterisation, and evaluation of a low-cost, non-wearable touchscreen interface integrating tactile feedback for surface manipulation tasks; and (3) a user study ($N=21$) evaluating response time, path-following accuracy, sense of telepresence, cognitive load, and usability across two feedback conditions.


\section{Related Work}
\label{sec:relatedwork}

Teleoperation of robotic systems in hazardous or inaccessible environments has been extensively studied across domains \cite{sheridan1992telerobotics}. Interface design plays a central role in enabling precise and cognitively efficient control \cite{hokayem2006bilateral}. Conventional joystick and button-based interfaces, while widely deployed, lack the spatial intuitiveness required for contact-rich manipulation. Touchscreen interfaces have emerged as a more natural alternative, offering direct mapping between finger motion and end-effector position. A prior study by the authors demonstrated that a touchscreen interface improved task efficiency and path-tracking accuracy compared to a joystick while reducing cognitive load \cite{garcia2026touchscreen}. However, the absence of haptic feedback was identified as a key limitation for maintaining operator situational awareness \cite{kenan2025robot}. The present work directly addresses this limitation.

Haptic feedback is essential for effective teleoperation of contact-rich tasks, as it conveys robot-environment interaction forces not easily communicated through vision alone \cite{okamura2004methods}. Kinesthetic feedback, as the most established modality reflects forces and torques through exoskeletons, manipulators, or haptic devices. However, it is prone to instability when interacting with stiff environments in bilateral teleoperation, and performance degrades significantly under communication delays \cite{lawrence1992stability}. Sensory substitution through visual cue-based force displays avoids these issues but imposes additional cognitive load by requiring the operator to monitor and interpret a separate information channel \cite{koritnik2010comparison}. Tactile feedback, by contrast, conveys force information through the same sensory modality as the physical interaction, reducing the need for cognitive conversion and enabling more intuitive perception of remote contact \cite{pacchierotti2017wearable}.

Electrovibration was first described by Mallinckrodt et al.~in 1953 \cite{mallinckrodt1953perception} and has since been revisited as a practical method for rendering tactile sensations on touchscreen surfaces \cite{bau2010teslatouch}. It operates by applying an alternating voltage to a conductive screen surface, modulating the electrostatic friction force perceived at the fingertip. This distinguishes it from vibrotactile actuators and piezoelectric elements, which require physical actuation and are difficult to scale to large screen surfaces, and which can cause fatigue or discomfort during prolonged use. The perceptual characteristics of electrovibration, including the effects of waveform, frequency, and amplitude on perceived roughness and texture, have been studied in detail \cite{vardar2017roughness}, and a range of distinguishable tactile sensations has been demonstrated \cite{sadia2022exploration}. 

Despite this progress, electrovibration has primarily been investigated for consumer touchscreen interaction and has not previously been applied to robot teleoperation as a real-time force-reflection channel. This represents a missed opportunity: by integrating electrovibration feedback with tactile robotic perception sensors on the end effector, it becomes possible to reflect remote surface interactions directly to the operator's fingertip with minimal latency, offering a practical route toward tactile telepresence \cite{lepora2026tactile}. This is particularly relevant to the tactile internet, where low latency is a key requirement for transparent force reflection and remains an open research challenge \cite{fettweis2014tactile, simsek20165g}.


\section{Methodology}
\label{sec:Methodology}

To evaluate the proposed interface for touchscreen-based robot teleoperation, a comparative user study was conducted against an industry-standard visual force representation. The objective was to conduct characterisation experiments to evaluate interface performance and user evaluations to assess usability, sense of telepresence, and cognitive load. Tasks were modelled on realistic scenarios, including responding to feedback, following a feedback-guided path, performing swabbing tasks across multiple surfaces in industry settings, and avoiding obstacles, all designed to replicate precise, contact-based manipulation over a surface.

\subsection{Participants}

A total of 21 participants took part in the study, 15 male and 6 female. Participants were recruited via email. Their ages ranged from 19 to 37 years (\emph{M} = 27, \emph{SD} = 4.32).  All participants reported being right-handed and used their right hand to interact with the interface, though right-handedness was not a recruitment requirement and occurred by chance. Most participants reported some prior experience with robots: 9 reported having controlled a robot once or twice, 5 occasionally, and 2 regularly, while 5 participants reported having never controlled a robot before.

\subsection{Study Environment and Context}

\begin{figure}[h]
\centering
\includegraphics[width=1.00 \columnwidth]{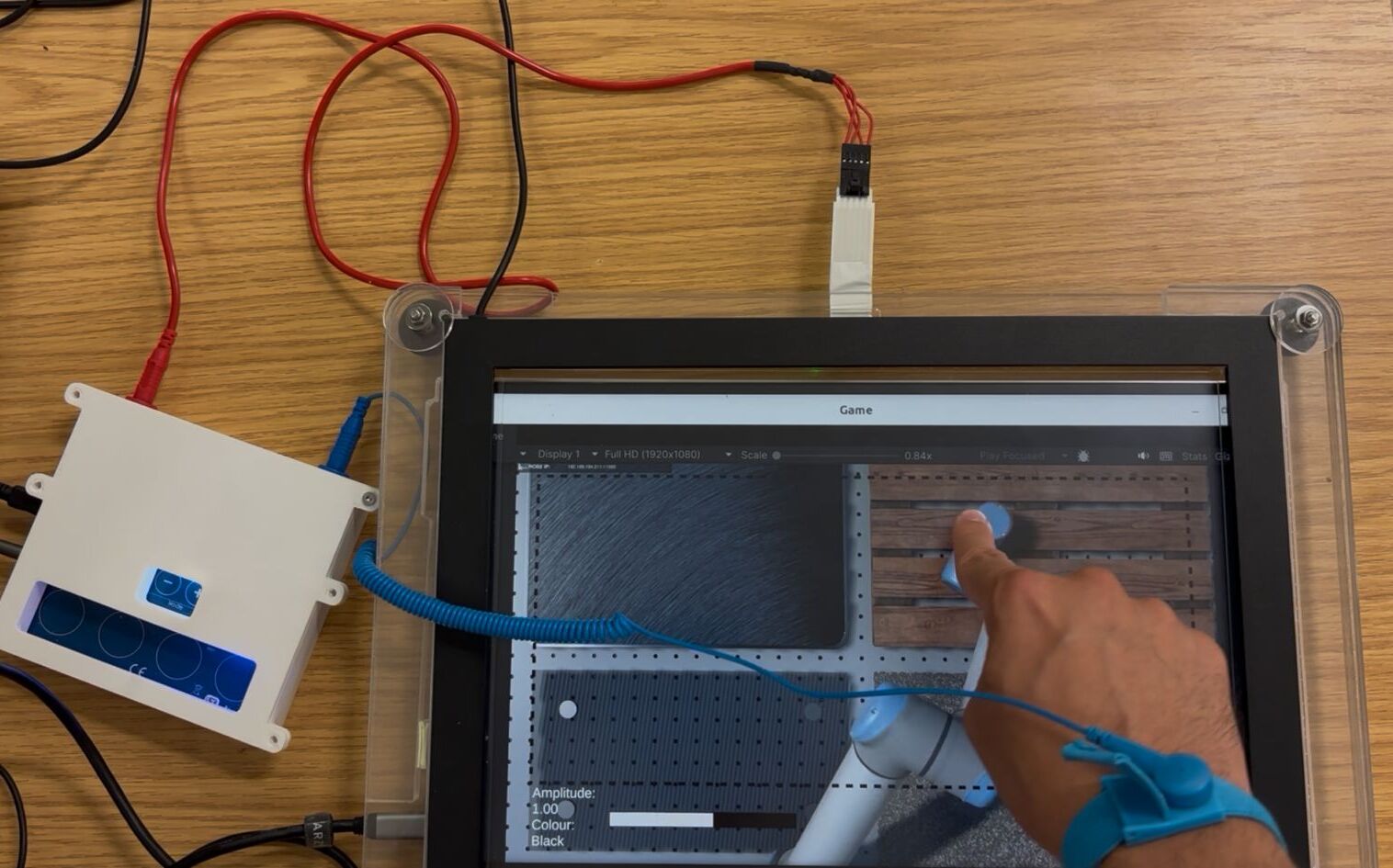}
\caption{An image representing the user study}
\label{fig:real}
\vspace{-15pt}
\end{figure}

The study took place at the Bristol Robotics Laboratory (Bristol, UK), where participants interacted with the interface, while the teleoperated robot and the environment were simulated in Unity (version 6.4)\footnote{Unity is a real-time 3D development platform used for simulation, visualisation, and interactive applications. See \url{https://unity.com/}.}. A simulated UR10 robotic arm was used throughout all experiments. Figure \ref{fig:real} shows a representative image of the user study.

\begin{figure}[ht]
    \centering
    \begin{subfigure}[b]{0.49\columnwidth}
        \centering
        \includegraphics[width=\linewidth]{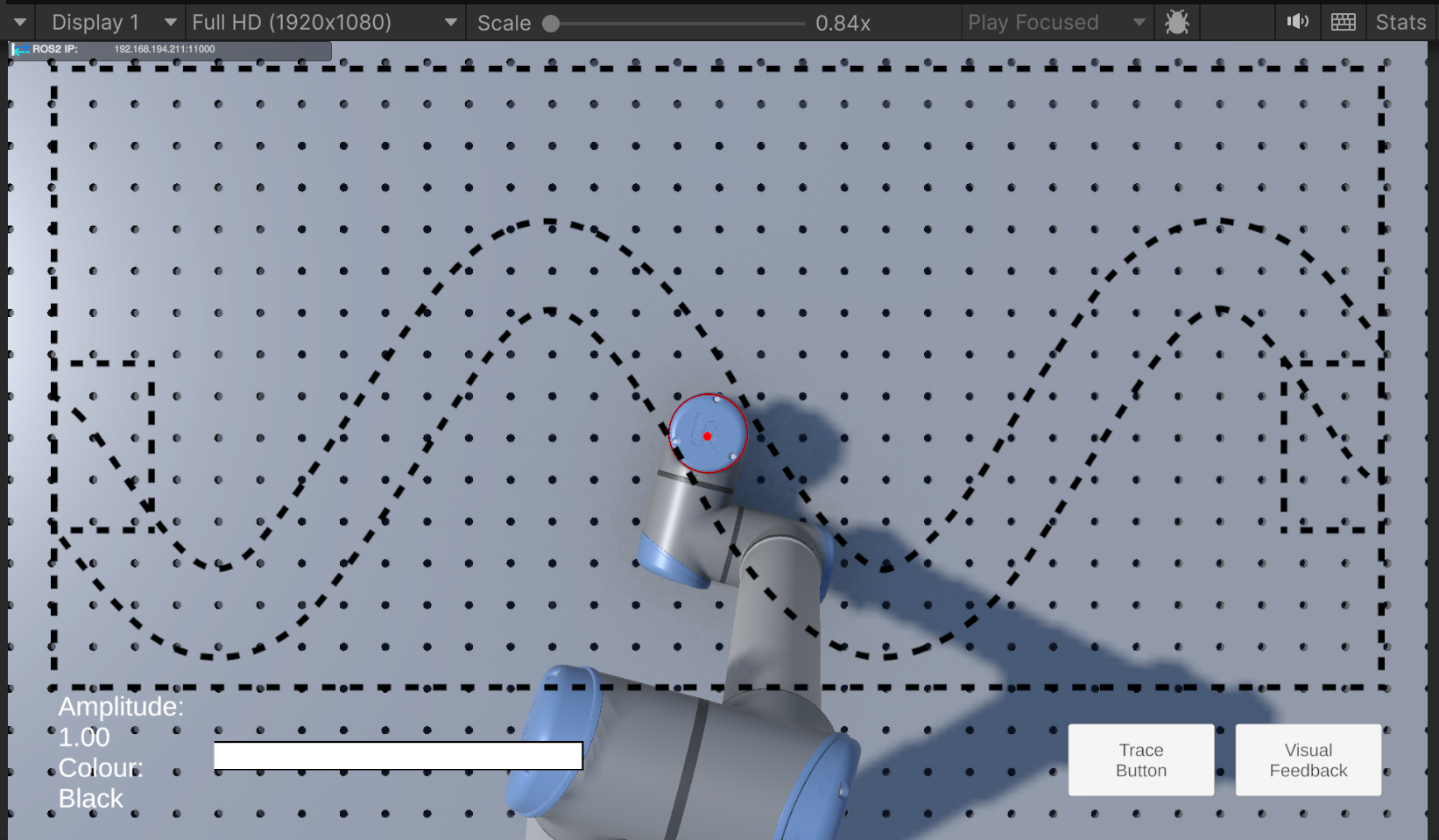}
        \caption{Path-following task}
        \label{fig:subfig1}
    \end{subfigure}
    \hfill
    \begin{subfigure}[b]{0.49\columnwidth}
        \centering
        \includegraphics[width=\linewidth]{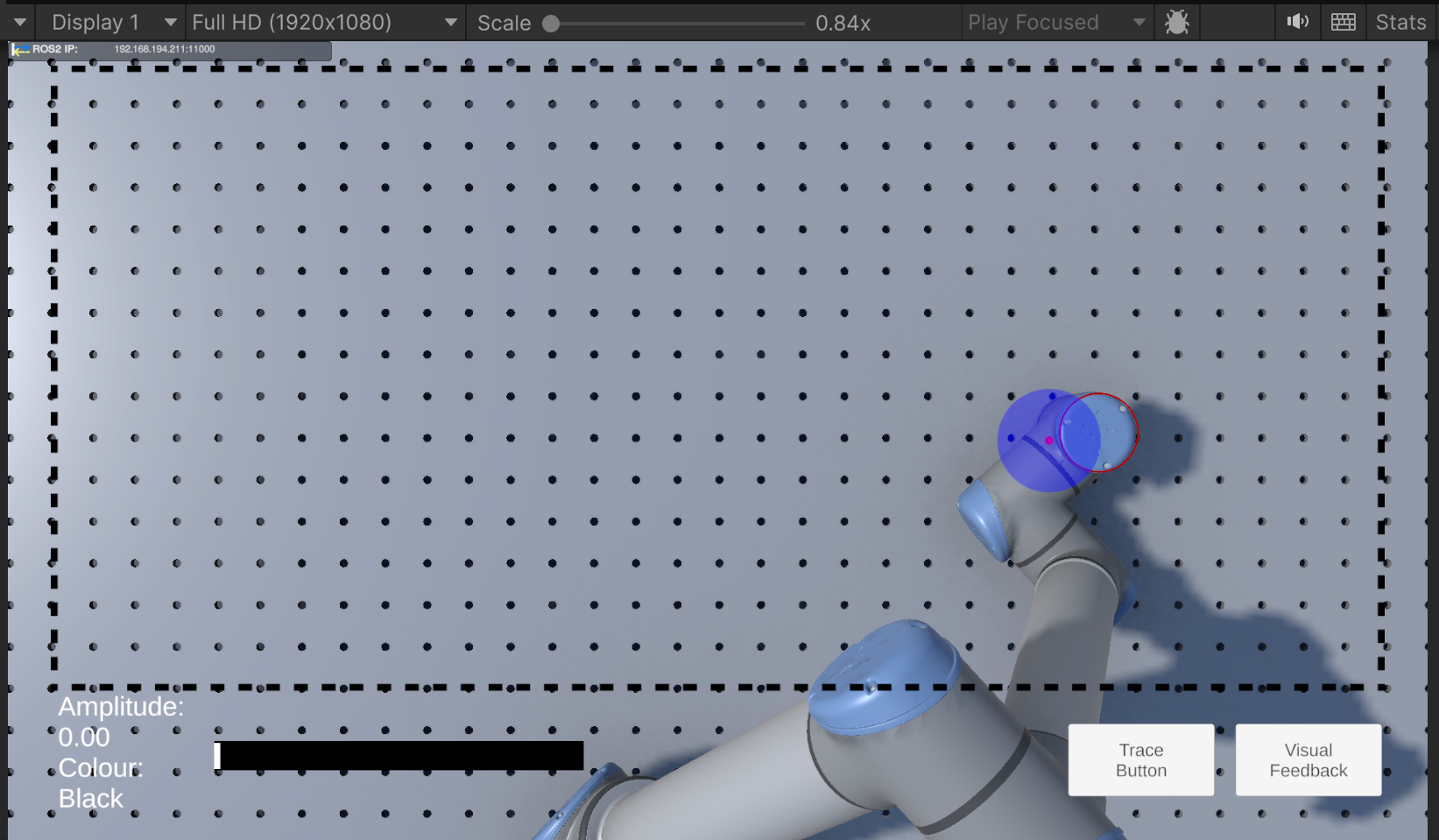}
        \caption{Response time task}
        \label{fig:subfig2}
    \end{subfigure}
    \caption{Two characterisation experiment tasks}
    \label{fig:characterisation}
    \vspace{-10pt}
\end{figure}

Teleoperation tasks were performed in two phases: characterisation tasks and user evaluation tasks, both tested under two conditions: (VF) visual feedback represented by a bar graph on the side of the screen without any haptic feedback, and (TF) tactile feedback without any visual feedback. All conditions used an identical camera feed. Characterisation experiments included two tasks: (i) path-following accuracy, and (ii) response time measurement. In the path-following task, a sine wave path was assigned for the robot end effector to follow which was not directly visible to participants, so they had to rely on visual or haptic feedback according to their assigned condition to follow this path. Two sine wave paths with varying amplitude and frequency were used to represent non-linear, non-rectangular trajectories, and path-following accuracy was recorded in terms of average distance and maximum distance per participant for both paths and both conditions. In the response time task, participants were asked to follow a blue circle on the screen with their finger and lift it upon receiving feedback, with response time recorded in milliseconds. Response time is recorded three times per participant for each condition. The two characterisation experiment tasks are illustrated in Fig. \ref{fig:characterisation}, showing the sine wave path used for path-following accuracy and the blue circle target used for response time measurement.

\begin{figure}[ht]
    \centering
    \begin{subfigure}[b]{0.48\columnwidth}
        \centering
        \includegraphics[width=\linewidth]{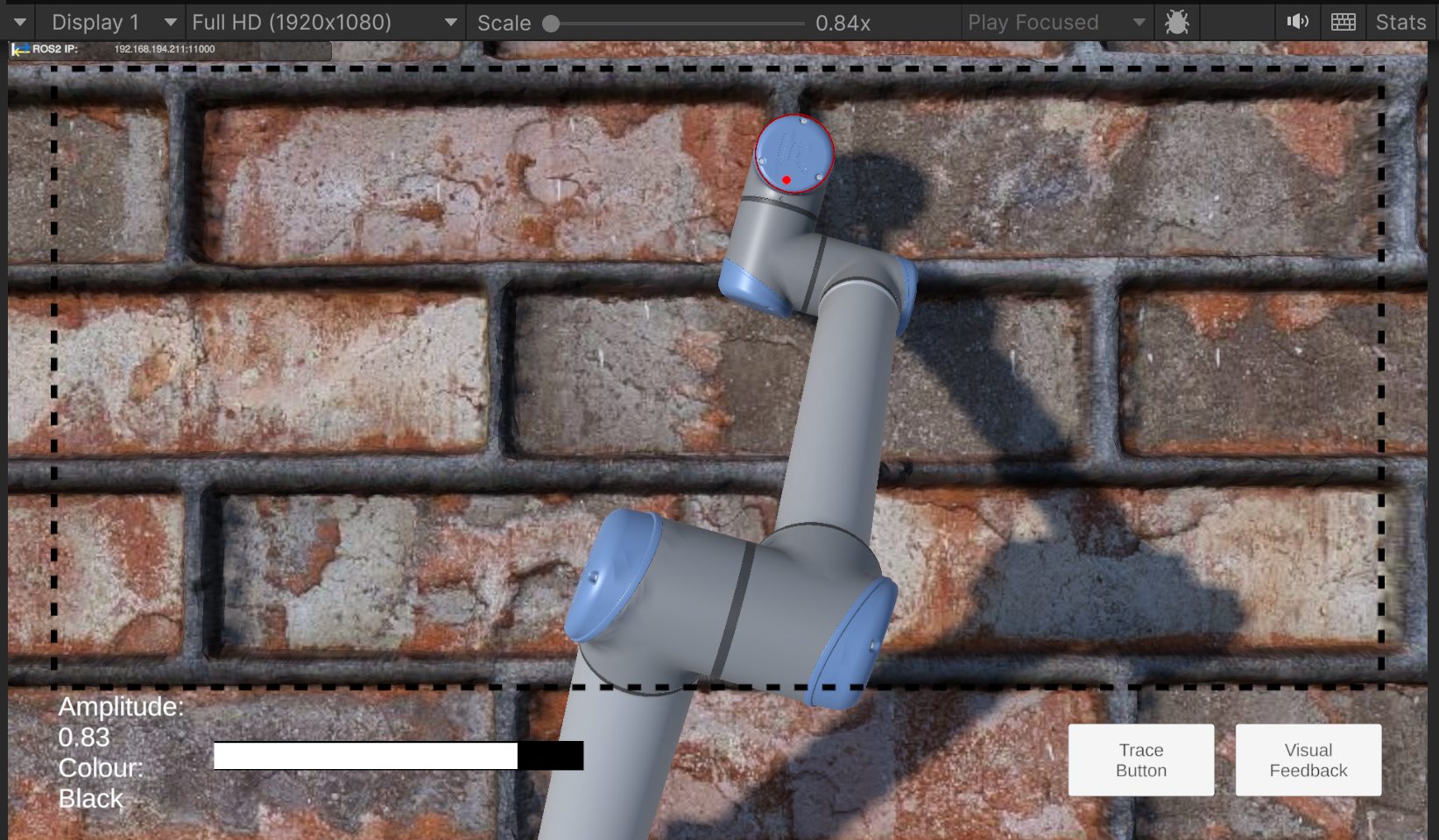}
        \caption{Defect Detection}
        \label{fig:subfig1}
    \end{subfigure}
    \hfill
    \begin{subfigure}[b]{0.48\columnwidth}
        \centering
        \includegraphics[width=\linewidth]{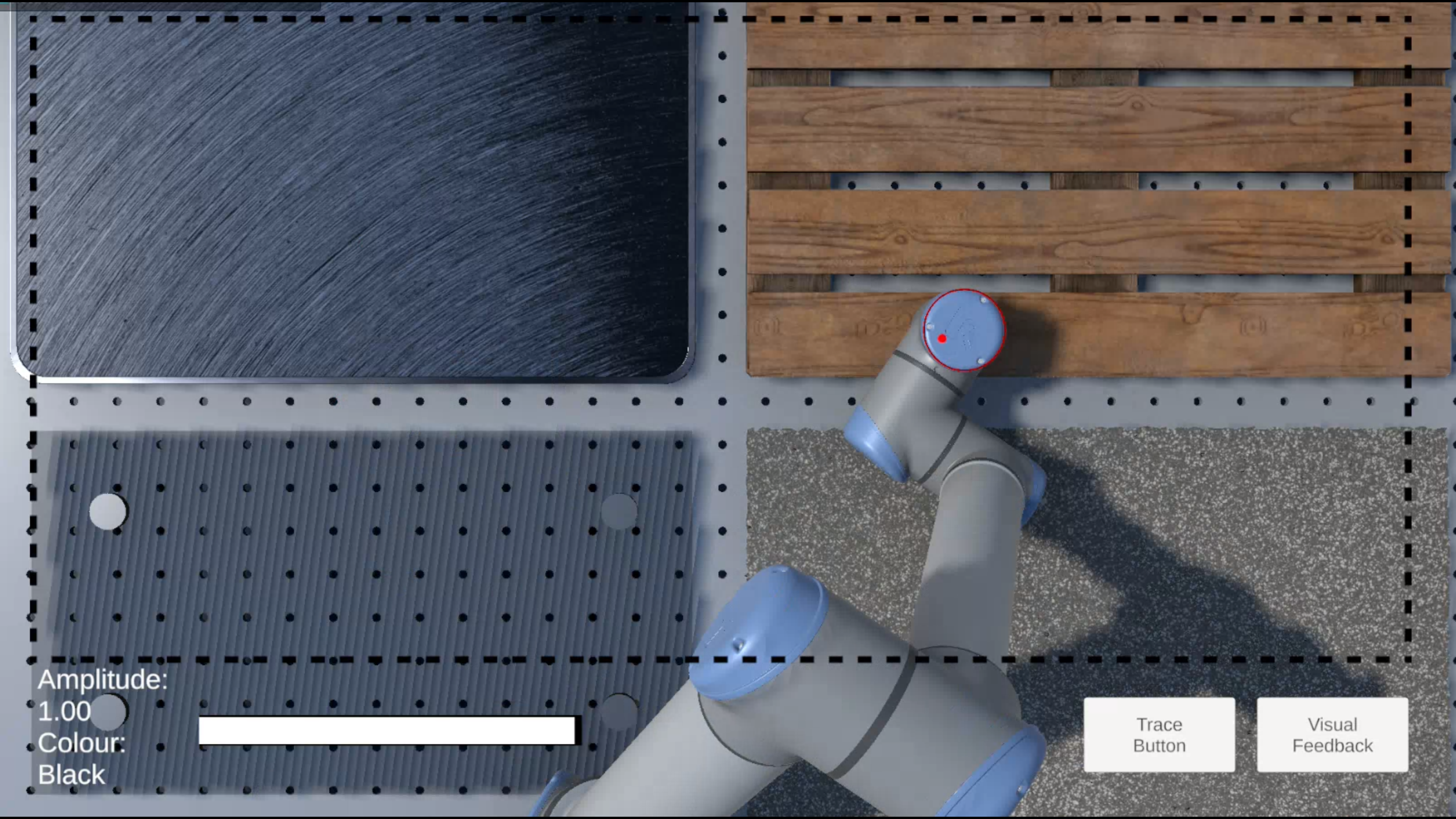}
        \caption{Multi-Surface Swabbing}
        \label{fig:subfig2}
    \end{subfigure}

    \vspace{4pt}

    \begin{subfigure}[b]{0.48\columnwidth}
        \centering
        \includegraphics[width=\linewidth]{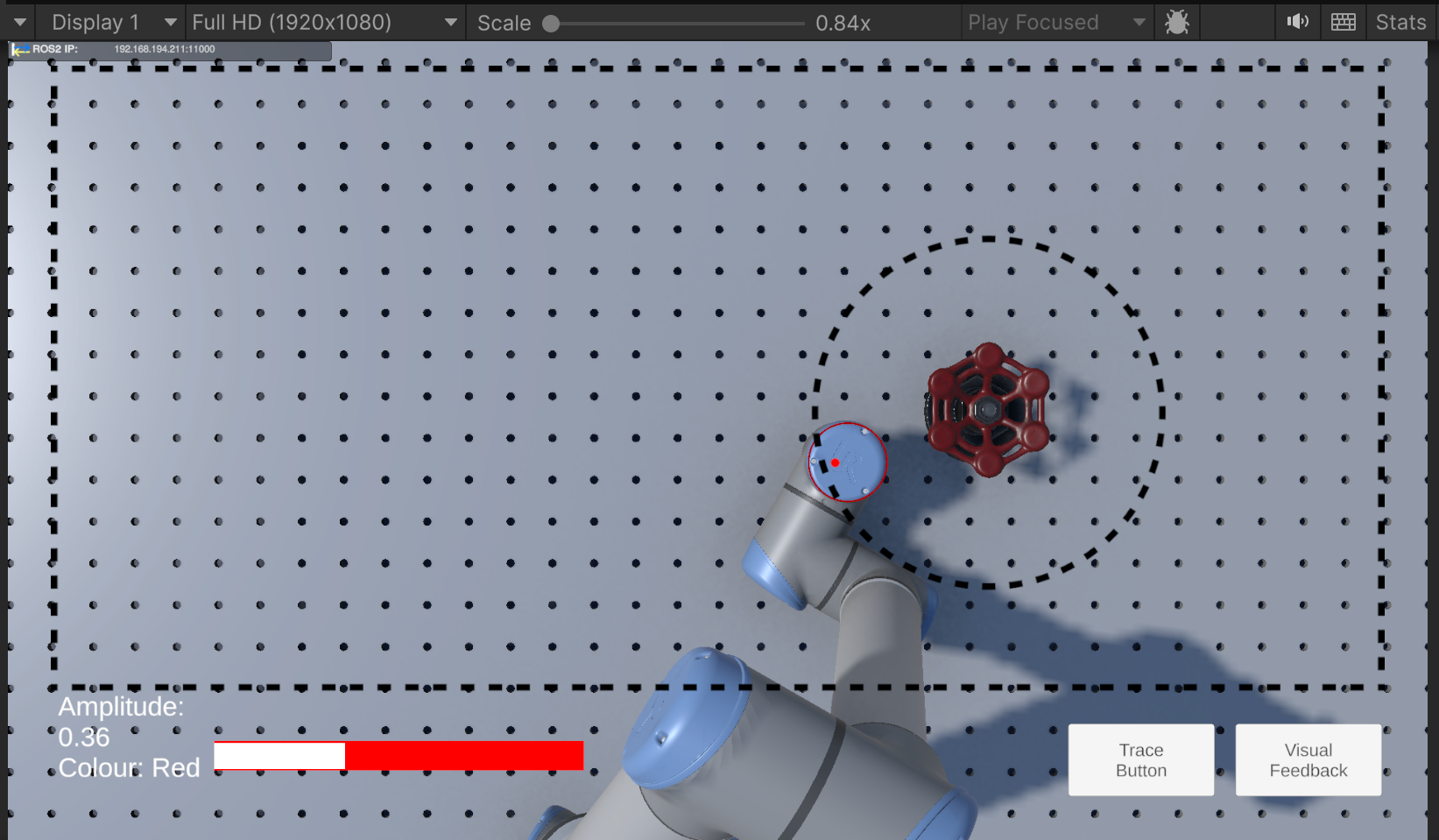}
        \caption{Obstacle Avoidance}
        \label{fig:subfig3}
    \end{subfigure}
    \hfill
    \begin{subfigure}[b]{0.48\columnwidth}
        \centering
        \includegraphics[width=\linewidth]{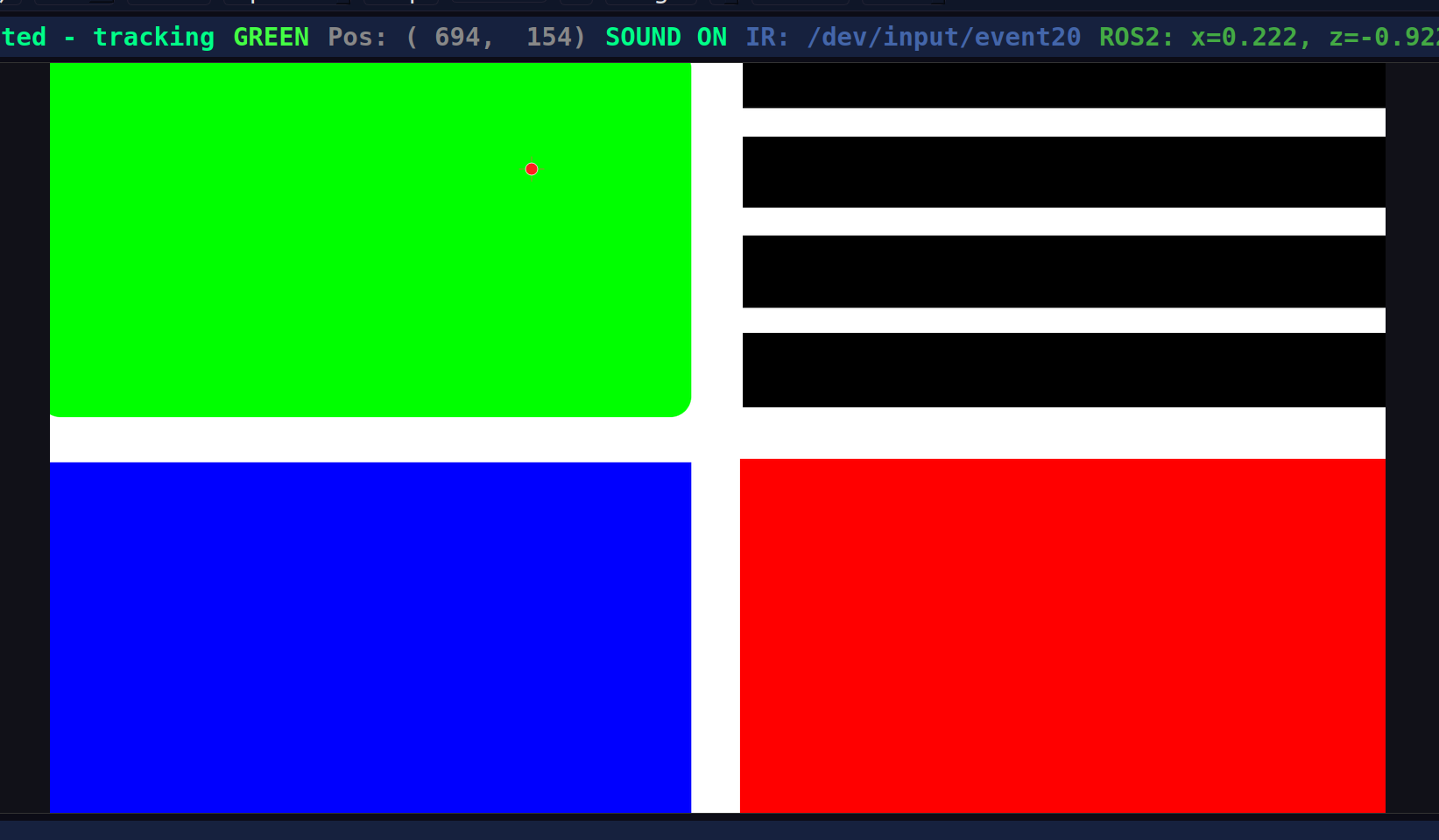}
        \caption{Tactile Feedback Map}
        \label{fig:subfig4_image}
    \end{subfigure}

    \caption{Overview of the evaluation tasks. In (d), different colours correspond to different textures presented to the participant in the multi-surface swabbing task, and white regions indicate no feedback.}
    \label{fig:evaluation_cases}
    \vspace{-10pt}
\end{figure}

For the user evaluation phase, there were three tasks: (i) surface defect detection, (ii) multi-surface swabbing task, and (iii) obstacle avoidance. In the first task, participants interacted with a simulated brick wall surface and received feedback when passing over defects between bricks, either as tactile feedback felt directly on their fingertips or as a visual bar graph representation, replicating the sensation of perceiving the surface directly. In the multi-surface scenario, four different surfaces provided distinct sensations through varying electrovibration frequency and amplitude, represented by different colours in the visual feedback condition. In the obstacle avoidance task, feedback acted as a collision warning system, with feedback amplitude increasing as proximity to the obstacle decreased.  The three user evaluation scenarios alongside the image processing for the multi-surface task are shown in Fig. \ref{fig:evaluation_cases}, depicting the brick wall defect detection, multi-surface swabbing, obstacle avoidance, and the corresponding visual image processing running in the background for multii-surface swabbing task.

\begin{figure}[ht]
    \centering
    \begin{subfigure}[b]{0.48\columnwidth}
        \centering
        \includegraphics[width=\linewidth]{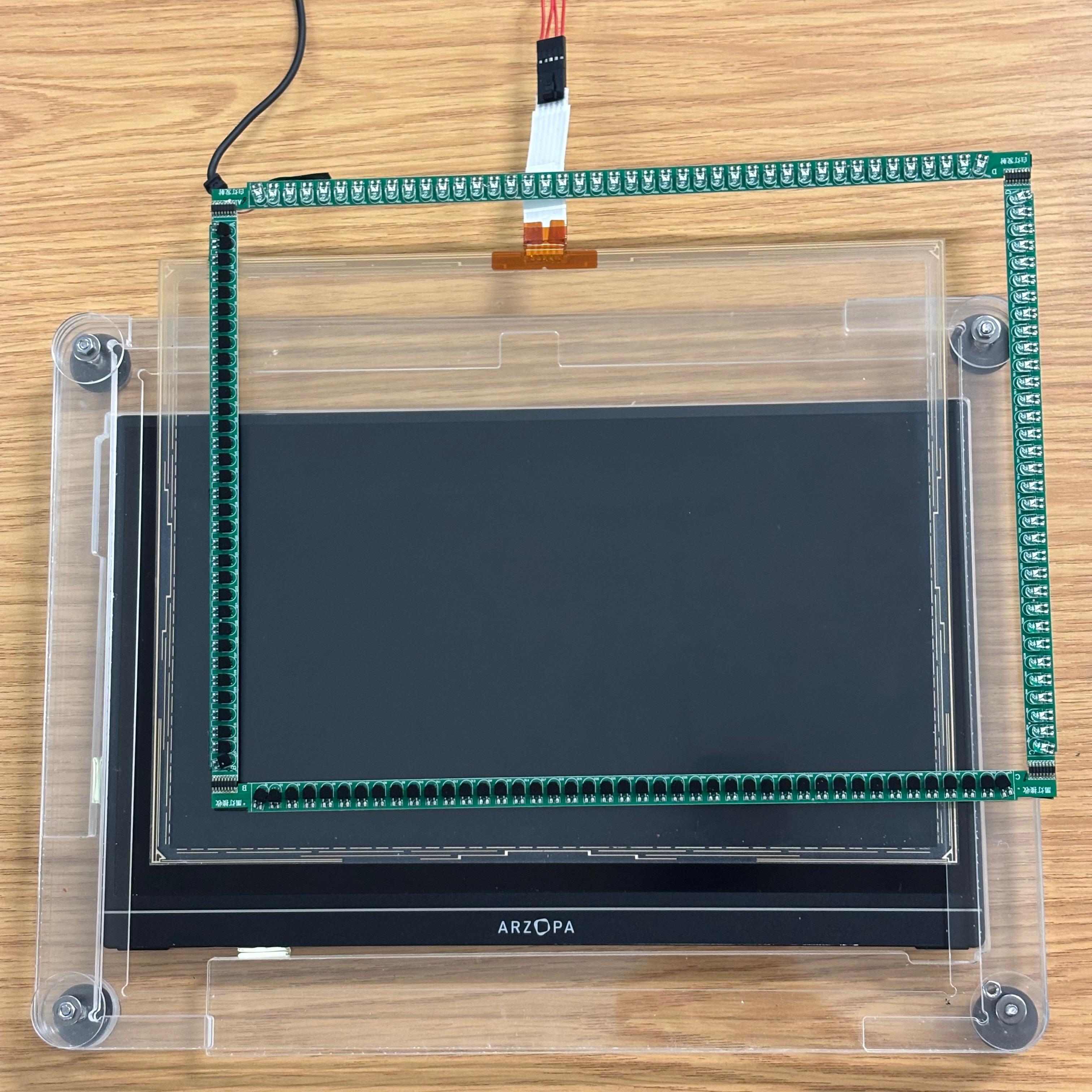}
        \caption{Different layers visible}
        \label{fig:subfig1}
    \end{subfigure}
    \hfill
    \begin{subfigure}[b]{0.48\columnwidth}
        \centering
        \includegraphics[width=\linewidth]{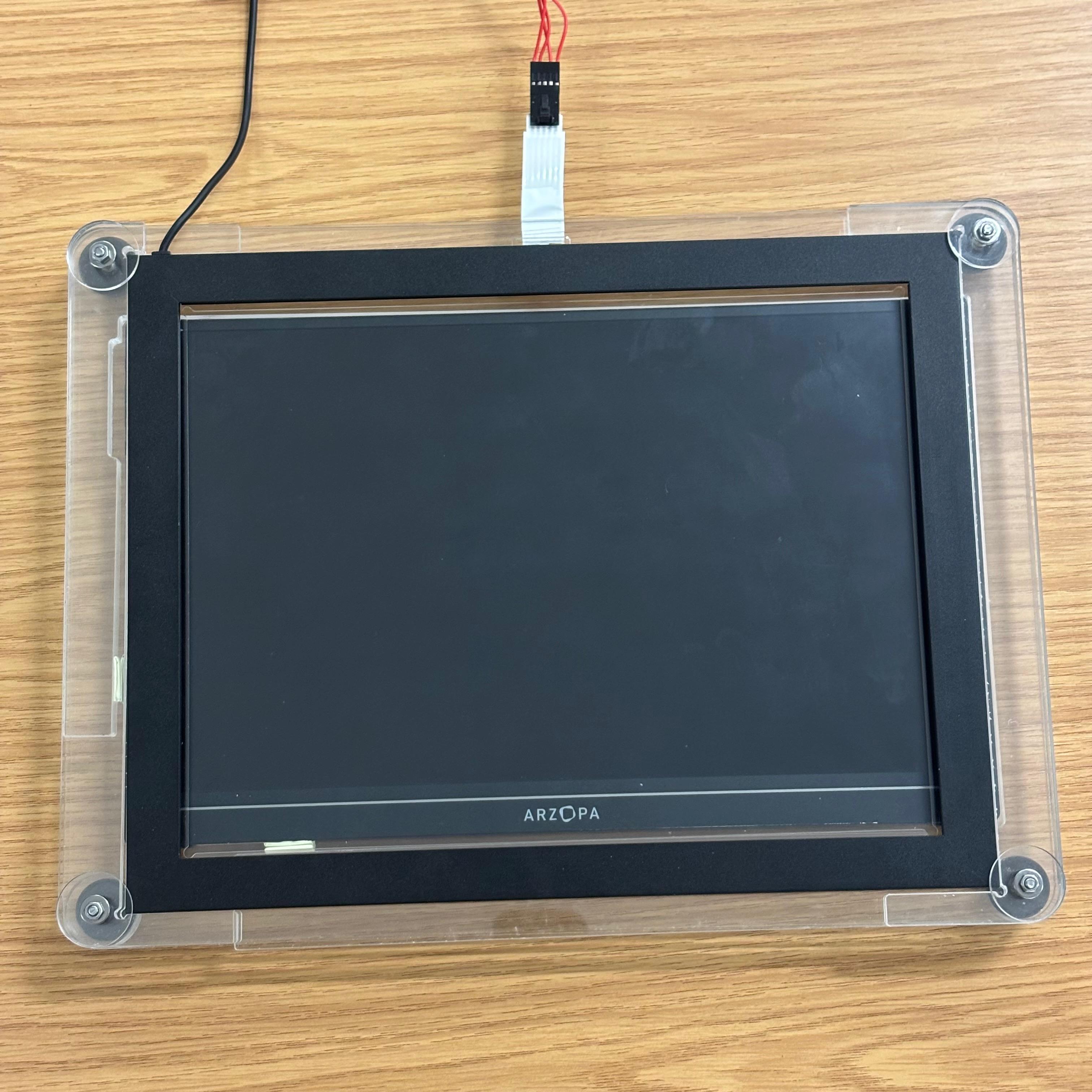}
        \caption{Closed format during study}
        \label{fig:subfig2}
    \end{subfigure}
    \caption{Touchscreen Interface}
    \label{fig:interface}
    \vspace{-10pt}
\end{figure}

As it is not feasible to apply high voltage directly to a commercial touchscreen, the interface was implemented using three separate layers. The bottom layer consisted of a monitor display (ARZOPA Portable Monitor, 15.6'', 1080P) providing visual output. A capacitive conductive touchscreen (3M 15-inch SCT3250) was placed in the middle layer to deliver the electrovibration effect. Since this layer did not track finger position, an IR touch screen frame (GreenTouch 15-inch) was placed on top for positional tracking. These three layers were integrated into a single interface using laser-cut acrylic components, enabling simultaneous visualisation, tactile feedback delivery, and finger position tracking. Details of the hardware used for generating and amplifying the electrovibration voltage are provided in Section \ref{sec:tactile_setup}. The three-layer touchscreen interface is shown in Fig. \ref{fig:interface}, with the layered components visible in the open configuration and the assembled unit as used during the study.

All 21 participants completed both phases under both conditions. During each run, the system recorded finger positions over the screen. Participants were given a practice trial for each phase and condition before proceeding. In all conditions, participants were instructed on the tasks they were expected to perform. The order of phases, conditions, task paths, and rendered environments was randomised across participants to minimise learning effects. To reduce bias, participants were not informed of the study's hypotheses. The open-source code for this project is publicly available.\footnote{The repository can be accessed at: \url{https://github.com/kenanalperen/Remote-Surfaces-at-Your-Fingertips.git}. It includes the list of hardware used, 3D models of the printed and laser-cut pieces, Python scripts used for the feedback, and the simulation environment used for controlling the robot, provided for reproducibility.}


\subsection{Subjective Measurements}

To assess cognitive workload, system usability, and changes in operator trust, both qualitative and quantitative data were collected. After each experimental condition, participants completed the NASA Task Load Index (TLX) \cite{nasatlx} to measure cognitive load. The System Usability Scale (SUS) questionnaire \cite{brooke1996sus} was also administered to evaluate the perceived usability of the interface. In addition, a task-specific questionnaire was administered to measure participants' sense of presence, comprising the following four statements rated on a 5-point Likert scale (1 = Strongly Disagree, 5 = Strongly Agree): (i) I felt present in the environment where the robot was operating. (ii) I felt as if I was controlling the robot directly rather than through a touchscreen. (iii) I felt a strong sense of control over the robot's actions. (iv) I felt engaged with the remote task while using the interface. All questionnaire responses were converted to a 100-point performance metric for consistency. In addition, open-ended suggestions for improving the interface feedback were collected from participants.

\subsection{Ethics}

This study was approved by the Research Ethics Committee of the University of the West of England, College of Arts, Technology \& Environment (Reference: 13470965). A participant information sheet was provided to all participants prior to their consent to take part, and signed consent forms were collected before the experiments started. Appropriate measures were taken to ensure participant confidentiality and data security throughout the study. Participants had the right to withdraw at any point and to request the removal of their data up to seven days following the study.

\section{Technical Details of the Telepresence Implementation}
\label{sec:technical_details}

\subsection{Tactile Feedback through Electrovibration}

Tactile feedback is conveyed to participants without notable delays through electrovibration, generated from the operator's finger position at the time of interaction. The mapping from interaction to feedback is determined by the end-effector's position relative to the surface, with the system classifying the surface region traversed and assigning a corresponding voltage amplitude, frequency, and waveform. Response-time measurements reflect the complete teleoperation loop, including simulation, processing, signal generation, communication, and hardware delays, as it is difficult to isolate electrovibration latency. These system-level delays were common to both conditions.

The effect is based on the well-established principle of a parallel-plate capacitor \cite{demarest1998engineering}. Its complete working principle remains an active research area, with multiple models attempting to accurately predict force values. A simplified model for the electrostatic normal force on the finger ($F_e$) modelled as an air-filled parallel-plate capacitor is given in Equation \ref{eq:electrostatic_force} \cite{shultz2015surface}, in terms of the contact area ($A$), the permittivity of free space ($\varepsilon_0$), the relative gap permittivity ($\varepsilon_g$), the gap voltage ($V_g$), and the gap separation ($d_g$). As the force output is proportional to the square of the applied voltage, the voltage sent to the screen serves as the control input for modulating the lateral friction force at the fingertip:

\begin{equation}
\label{eq:electrostatic_force}
F_e = \frac{A\varepsilon_0\varepsilon_g}{2} \left( \frac{V_g}{d_g} \right)^{2}
\end{equation}

\begin{figure}[h]
\centering
\includegraphics[width=1.00\columnwidth]{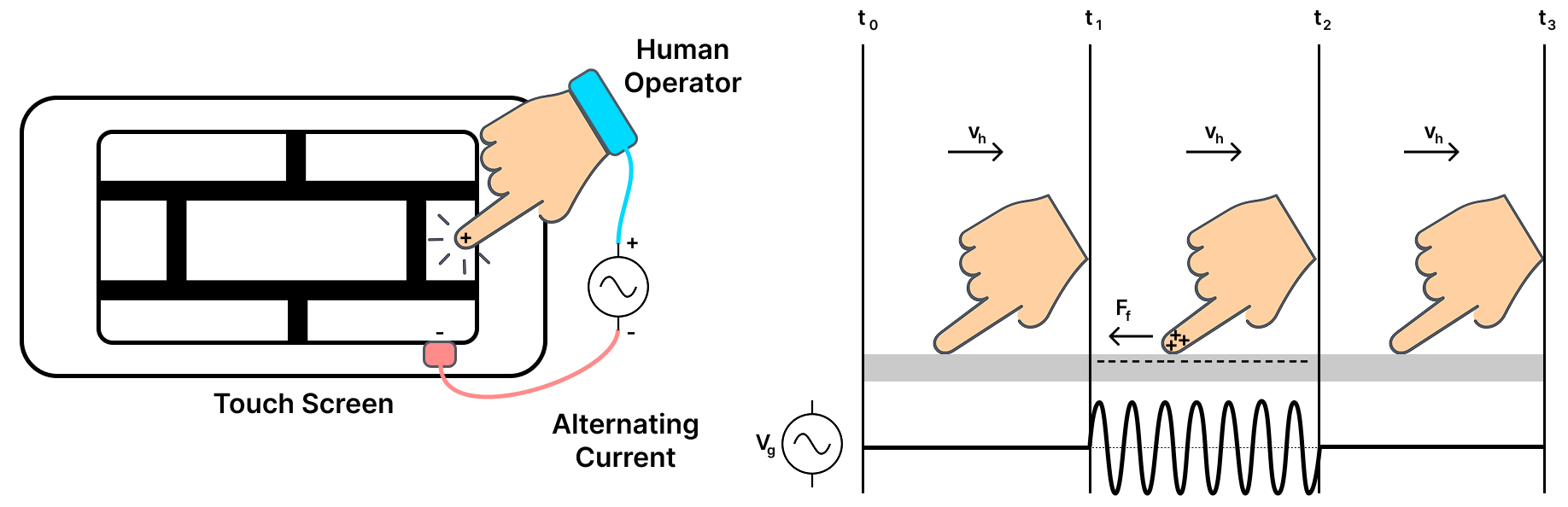}
\caption{Working principle illustration of electrovibration.}
\label{fig:details}
\vspace{-15pt}
\end{figure}

As the electrostatic normal force increases, the friction force experienced at the fingertip increases proportionally, following Coulomb's friction law ($F_f = \mu F_e$), where $F_f$ is the induced lateral friction force and $\mu$ is the coefficient of friction between the fingertip and the screen surface. Figure \ref{fig:details} illustrates the mechanics of the electrovibration setup, where $t$ represents timestamps, and $v_h$ is the velocity of the fingertip. When an alternating voltage is applied between $t_1$ and $t_2$, the accumulation of charge between the fingertip and the touchscreen induces an electrostatic force, resulting in additional transient friction. By timing the applied voltage to match surface defects recorded in a remote environment, the system replicates the tactile sensation of directly touching the remote environment itself.

Despite its name, electrovibration does not involve mechanical vibration unlike the vibrotactile actuators found in commercial smartphones. Because the proposed interface contains no mechanically moving parts, signal latency is inherently minimised. Furthermore, the sensation perceived at the fingertip is mechanical rather than electrical, as the current passing through the tissue remains well below the threshold for electrical stimulation. The tactile illusion arises from modulations of the friction force in the tangential plane.

\subsection{Tactile Feedback Experiment Setup}
\label{sec:tactile_setup}

For generating and amplifying the alternating electric voltage based on the analogue reference signal sent by the computer, a Texas Instruments DRV2667EVM-CT TouchPath Evaluation Module was used\footnote{Details about the hardware can be accessed at \url{https://www.ti.com/tool/DRV2667EVM-CT}.}. The analogue reference is first converted to an audio output signal, which is then conveyed to the hardware module via the audio jack input. The hardware was operated in Mode~0-B4.

\begin{figure}[h]
\centering
\includegraphics[width=1.00\columnwidth]{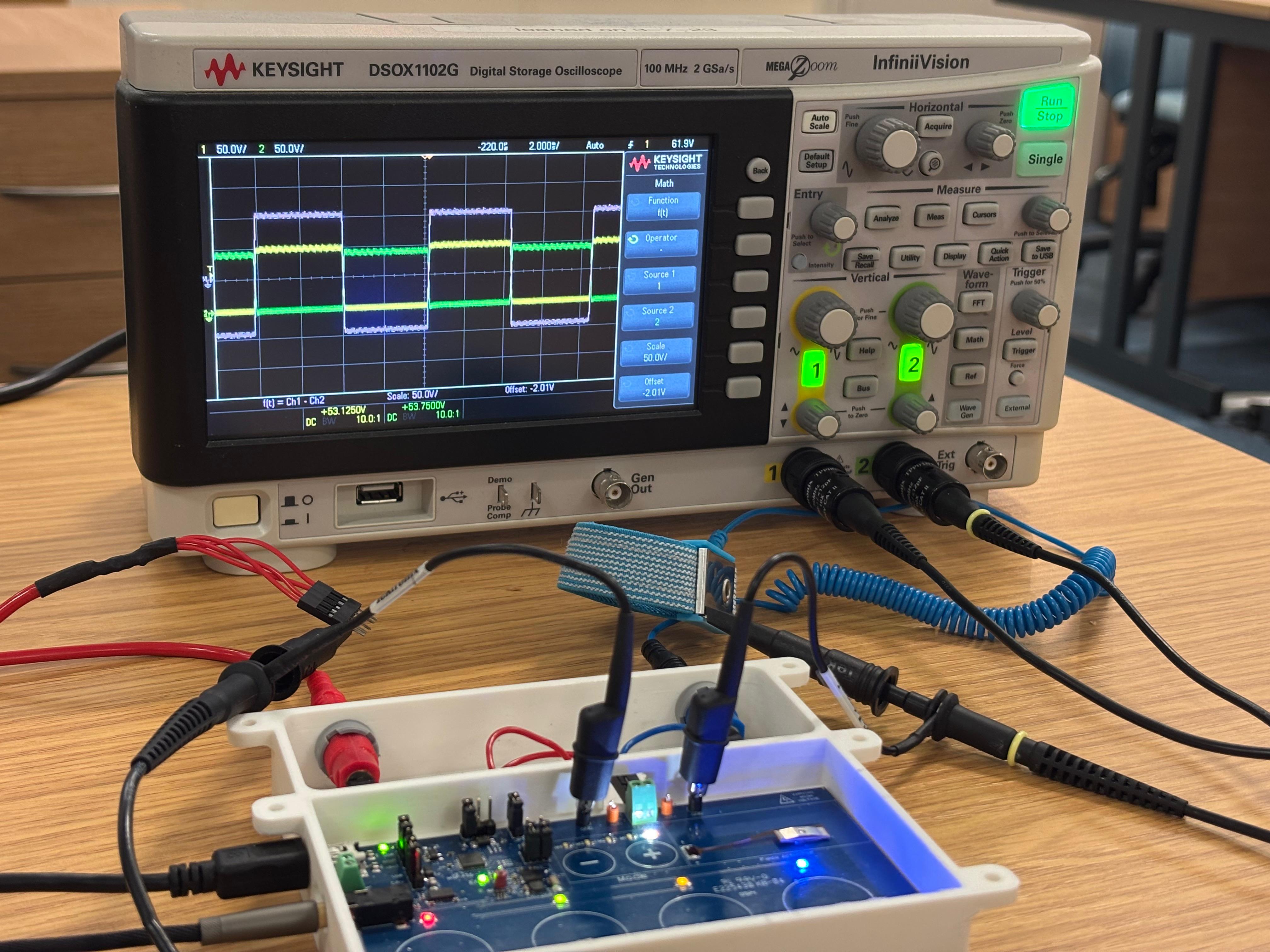}
\caption{Oscilloscope reading of the voltage difference output between two ports}
\label{fig:oscilloscope}
\vspace{-15pt}
\end{figure}

The tactile sensation depends on the amplitude, frequency, and wave shape of the voltage output. Based on the literature on maximally differentiable haptic sensations \cite{vardar2017roughness}, the primary signal used in this study was a square wave at 200~V\textsubscript{pp} and 120~Hz, while the multi-surface implementation used varying parameters across surfaces. An example square wave output at 200~V\textsubscript{pp} and 120~Hz is shown in Fig. \ref{fig:oscilloscope}, where the green and yellow lines represent the individual voltage outputs of ports, and the white line represents the voltage difference between them, obtained through the oscilloscope's math (minus) function.

\begin{figure}[ht]
    \centering
    \begin{subfigure}[b]{0.48\columnwidth}
        \centering
        \includegraphics[width=\linewidth]{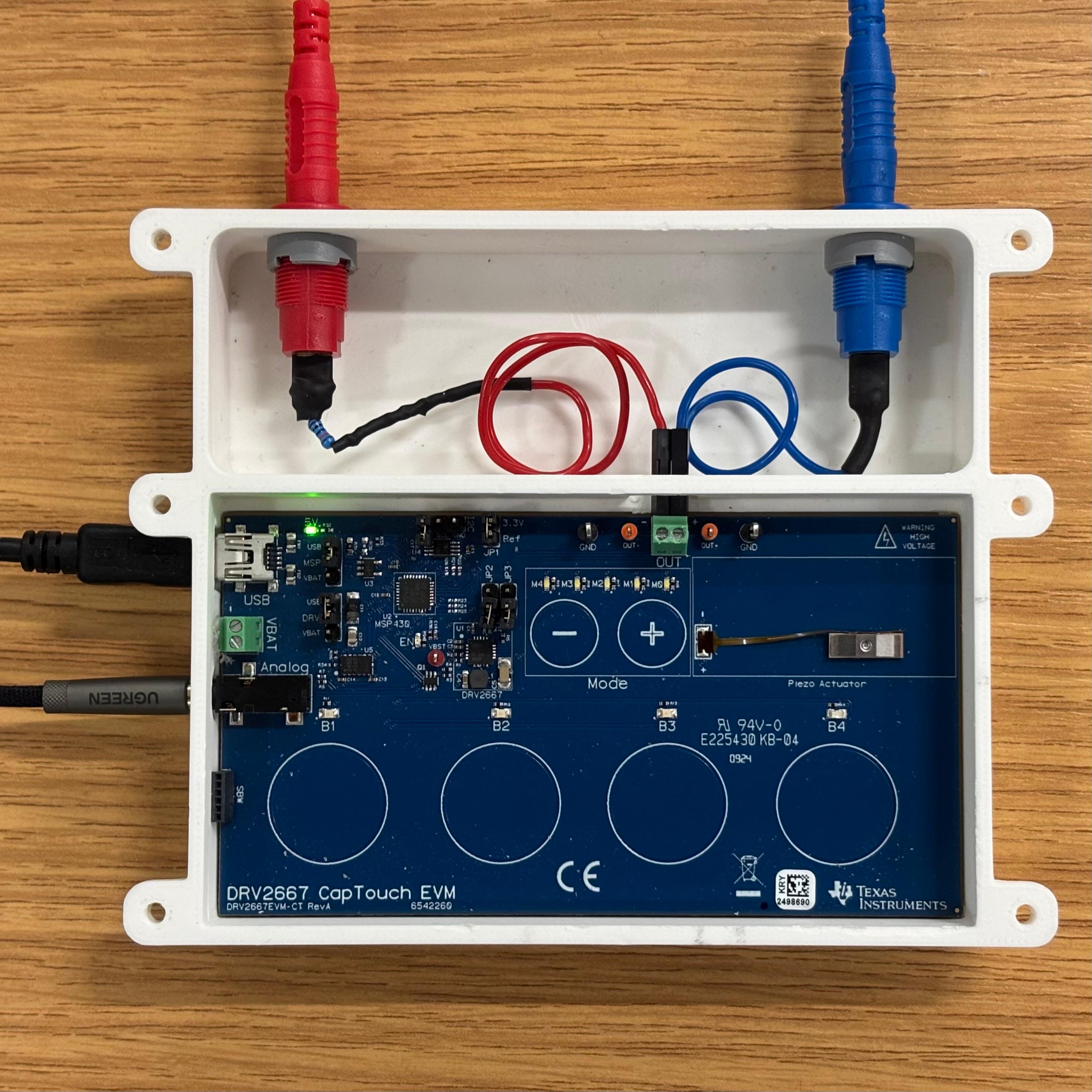}
        \caption{Inner components visible}
        \label{fig:subfig1}
    \end{subfigure}
    \hfill
    \begin{subfigure}[b]{0.48\columnwidth}
        \centering
        \includegraphics[width=\linewidth]{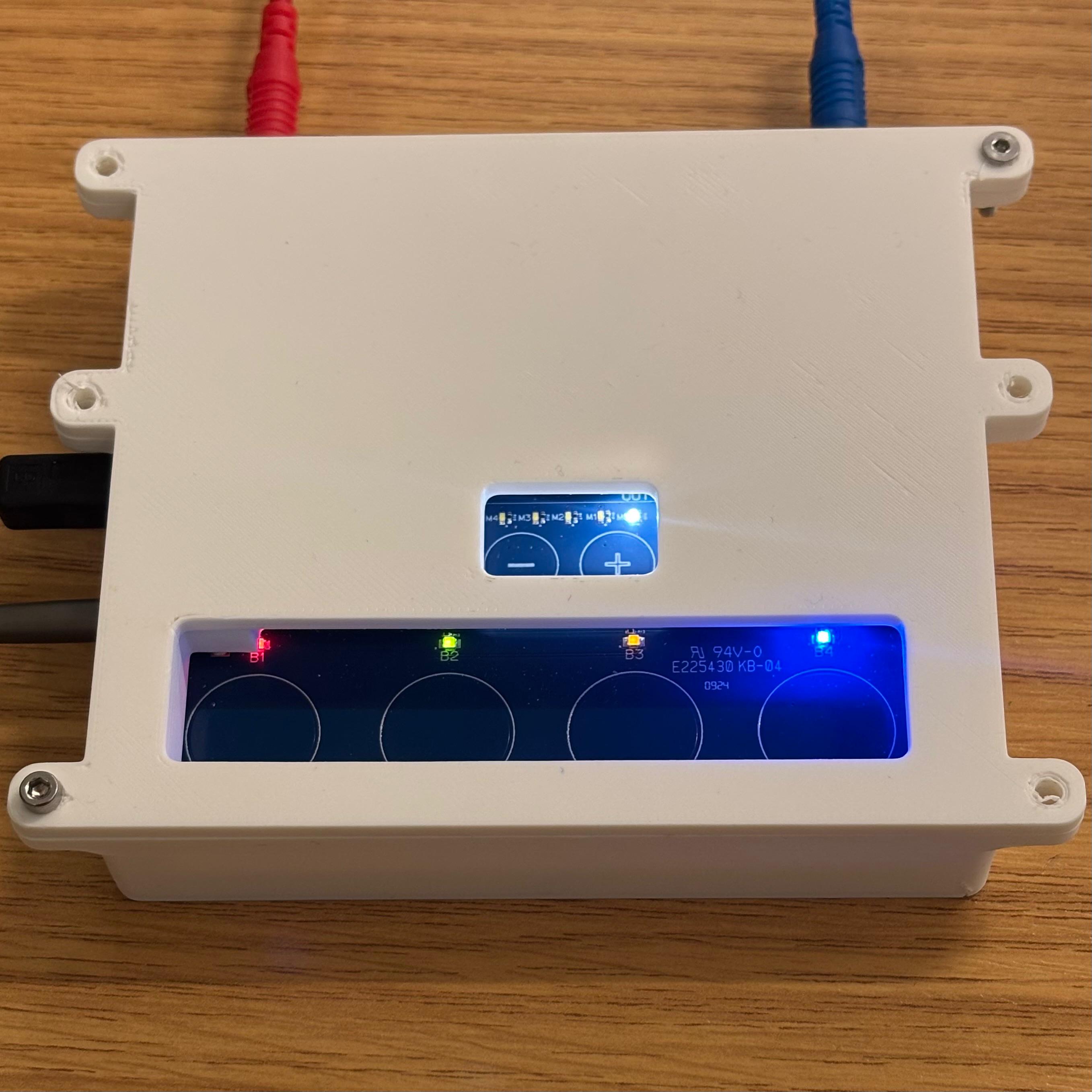}
        \caption{Closed format during study}
        \label{fig:subfig2}
    \end{subfigure}
    \caption{Hardware used to generate and amplify voltage signals to the screen.}
    \label{fig:hardware}
    \vspace{-10pt}
\end{figure}

The system operates at high voltages of up to 200~V\textsubscript{pp}, which can be hazardous if not handled correctly. Although the power supply draws from a computer USB port, a short circuit between the differential voltage outputs could still be dangerous. To ensure safe operation, an enclosed isolator box was constructed, exposing only the touch pads and enclosing all power outputs during experiments. An additional 100~k$\Omega$ resistor was placed on the output line to the touchscreen, limiting the current to 1~mA in the event of an unexpected short circuit, within safe limits. All cable connectors used shrouded banana sockets to prevent unintended contact with live terminals. The antistatic wristband incorporated a built-in 1~M$\Omega$ resistor, reducing the current through the operator's fingertip to approximately 0.2~mA, well below established safety thresholds. The hardware enclosure used during the user studies is shown in Fig. \ref{fig:hardware}.

\section{Hypotheses}
\label{sec:Hypothesis}

Building on research in tactile perception and collaborative HRI, four different hypotheses are formulated:

\begin{itemize}

    \item [\textbf{H1}] Electrovibration-based tactile feedback can convey force outputs without notable delays and will be perceivable by participants.

    \item [\textbf{H2}] Participants in condition TF (tactile feedback) will demonstrate faster response times compared to participants in condition VF (visual cue-based force feedback).

    \item [\textbf{H3}] Participants in condition TF will report a significantly higher sense of telepresence compared to condition VF.

    \item [\textbf{H4}] Participants in condition TF will report significantly lower cognitive workload compared to condition VF.
    
\end{itemize}

These hypotheses are motivated by prior research in tactile perception, HRI and human factors.

\textbf{H1} is grounded in findings \cite{bau2010teslatouch, sadia2022exploration} showing that electrovibration is capable of rendering tactile feedback under the optimal operating conditions described in \cite{vardar2017roughness}. To further validate, participants will be asked whether they perceive the tactile feedback clearly.

\textbf{H2} is grounded in the established difference in neural processing between tactile and visual stimuli \cite{ng2012finger, kim2020visual}, where tactile reaction times have been shown to be 34\% shorter than visual reaction times, with tactile responses averaging approximately 241 ms compared to 329 ms for visual stimuli. As TF delivers feedback directly to the fingertip while VF requires visual processing and a planned motor response, participants in TF are expected to demonstrate faster response times.

\textbf{H3} is grounded in the principle that matching the feedback modality to the physical interaction strengthens the sense of presence \cite{wagener2022influence, cooper2018effects}. VF acts as a sensory substitute requiring cognitive conversion, whereas in TF the operator receives force information directly on the fingertip, matching the natural modality of touch and improving spatial presence and involvement.

\textbf{H4} is motivated by evidence that monitoring interaction forces through visual displays increases cognitive workload by dividing operator attention \cite{orun2019effect}. While visual feedback can improve performance \cite{horeman2012visual}, direct force feedback provides greater benefits and may enhance telepresence \cite{currie2017role, oppici2023does}. Therefore, participants in TF are expected to report lower cognitive workload than those in VF. To evaluate this, participants completed a questionnaire assessing cognitive load, interface usability, and sense of telepresence.

\section{Results}
\label{sec:Results}
The user study comprised two phases. The first phase aimed to measure task performance, while the second assessed operator perception of the proposed teleoperation interface with tactile feedback. All 21 participants completed the study and their data are included in both phases.

\subsection{Characterisation Experiment Results}

For each task, trial-level measurements were averaged across trials for each participant and condition prior to statistical analysis (2 trials per condition for the path-following task and 3 trials per condition for the response-time task), yielding $n=21$ paired samples for all tests reported below. 

For path-following accuracy, condition VF (visual cue-based force feedback) had an average position error of 65.96 pixels per participant ($SD=21.24$, $Min=42.63$, $Max=123.78$), while condition TF (tactile feedback) had an average of 63.84 pixels ($SD=19.54$, $Min=35.22$, $Max=126.19$). A paired $t$-test revealed no significant difference between conditions ($p=.76$, $t=0.31$, $d=0.07$). Similarly, the average maximum error was 228.20 pixels ($SD=79.96$, $Min=125.66$, $Max=411.62$) for VF and 226.00 pixels ($SD=73.24$, $Min=133.08$, $Max=438.84$) for TF, again with no significant difference ($p=.92$, $t=0.10$, $d=0.02$). 

\begin{figure}[h]
\centering
\includegraphics[width=1.00\columnwidth]{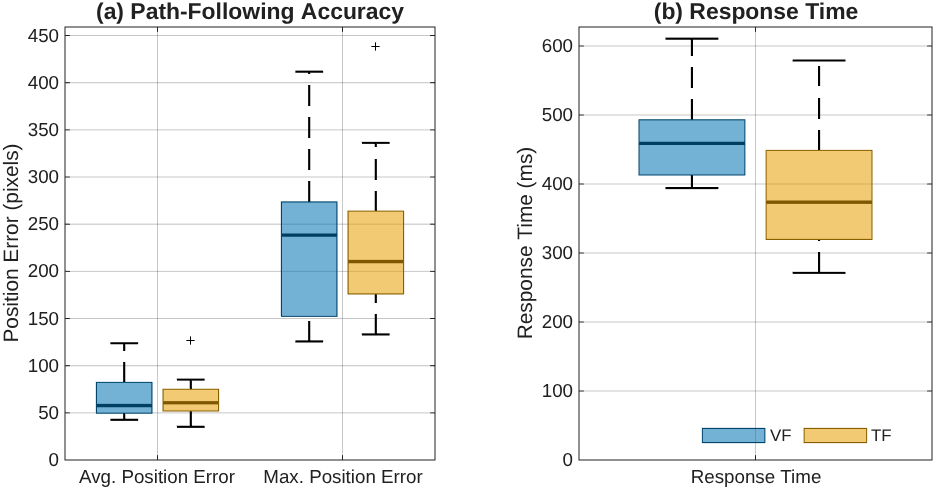}
\caption{Characterisation results shown in boxplots.}
\label{fig:results_char}
\vspace{-15pt}
\end{figure}

For the response time task, VF had an average response time of 466.92 ms ($SD=62.11$, $Min=394.12$, $Max=610.54$), while TF had an average of 395.24 ms ($SD=92.82$, $Min=271.18$, $Max=578.94$), a reduction of 71.68 ms (15.35\%). This difference was statistically significant ($p=.0016$, $t=3.64$, $d=0.79$). Characterisation experiment results are visualised as boxplots in Figure \ref{fig:results_char}.

\subsection{User Evaluation Results}

Participants responded to the questionnaire items based on their overall experience with each condition rather than on individual tasks. Each condition included 21 samples. All questionnaire scores were normalised to a 100-point scale. For SUS and telepresence, higher scores indicate greater usability and stronger sense of presence, respectively. For NASA-TLX, higher scores indicate higher cognitive load.

In terms of cognitive load, measured by NASA-TLX scores, VF had an average of 24.78 ($SD=19.91$), with the lowest sub-scale score for mental demand ($M=31.35$), while TF had an average of 27.04 ($SD=20.82$), with the lowest sub-scale score for effort ($M=33.89$). A paired $t$-test revealed no significant difference between conditions ($p=.57$, $t=0.58$, $d=0.13$).

In terms of usability, measured by SUS scores, VF had an average of 79.88 ($SD=13.03$), with the lowest sub-scale score for frequency of use ($M=53.45$), while TF had an average of 78.39 ($SD=17.24$), with the lowest sub-scale score again for frequency of use ($M=70.12$). A paired $t$-test revealed no significant difference ($p=.60$, $t=0.53$, $d=0.12$).

\begin{figure}[h]
\centering
\includegraphics[width=1.00\columnwidth]{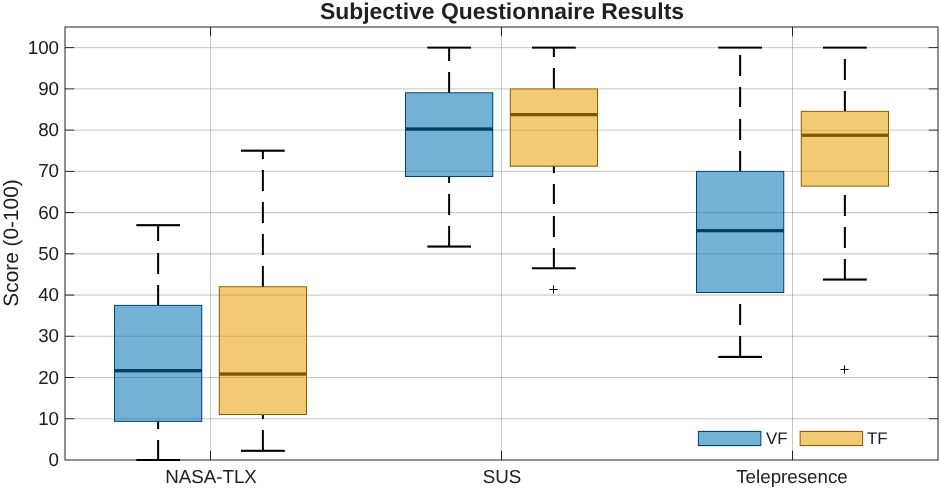}
\caption{User evaluation results shown in boxplots.}
\label{fig:results_user_ev}
\vspace{-15pt}
\end{figure}

In terms of sense of telepresence, VF had an average of 56.36 ($SD=21.00$), with the lowest sub-scale score for feeling direct control of the robot ($M=38.21$), while TF had an average of 73.87 ($SD=18.43$), with the lowest sub-scale score again for feeling direct control ($M=68.33$). The tactile feedback condition yielded an increase of 17.5 points (31\%) in average telepresence score, which was statistically significant ($p<0.001$, $t=-4.11$, $d=0.90$). User evaluation results are visualised as boxplots in Figure \ref{fig:results_user_ev}.

Participants were also asked to rank the perceived importance of the interface for each of the three evaluation tasks within each condition, where a rank of 1 indicated the most important task and a rank of 3 the least important task: (1) surface defect detection, (2) multi-surface swabbing, and (3) obstacle avoidance. For VF, the average rankings were 2.24, 2.10, and 1.67 ($SD=0.62$, $0.94$, $0.80$), respectively, while for TF they were 1.86, 1.71, and 2.43 ($SD=0.73$, $0.84$, $0.75$). A two-way repeated-measures ANOVA showed no significant difference in rankings across tasks for VF ($F(2,40)=1.94$, $p=.157$). For TF, a significant difference was observed ($F(2,40)=3.33$, $p=.046$).

\section{Discussion}
\label{sec:Discussion}

The experiment results broadly align with the proposed hypotheses, with strong 
support for H1, H2, and H3, while H4 requires further assessment.

\textbf{H1} is supported by all 21 participants reporting successful tactile perception and meaningfully engaging with both characterisation tasks, confirming that the interface can convey force outputs with sufficiently low latency for teleoperation applications.

\textbf{H2} is supported by the response time results, where participants in TF (tactile feedback) responded on average 71.68ms faster than in VF (visual feedback) (15.35\% reduction), with a statistically significant difference ($p<0.001$, $d=0.96$). This is consistent with the established advantage of tactile over visual neural processing pathways \cite{ng2012finger, kim2020visual}, confirming that delivering force feedback directly to the fingertip reduces operator response latency.

\textbf{H3} is strongly supported by the telepresence results, where TF yielded a 31\% higher sense of telepresence than VF ($p<0.001$, $d=0.90$). The largest gap between conditions was observed for feeling direct control of the robot ($M=38.21$ for VF vs. $M=68.33$ for TF), suggesting that matching the feedback modality to the natural modality of touch is particularly effective at enhancing telepresence \cite{wagener2022influence, cooper2018effects}.

\textbf{H4} requires further assessment, as NASA-TLX did no show a significant difference between conditions ($p=.57$, $d=0.13$ and $p=.60$, $d=0.12$, respectively). A possible explanation is that while VF required dividing attention to monitor a visual bar graph, TF required interpreting unfamiliar electrovibration cues, resulting in comparable cognitive demands overall for participants encountering this feedback modality for the first time. However, most participants indicated a preference for the tactile feedback condition in their open-ended responses, and suggested that combining both feedback modalities could yield the best overall performance.

The task ranking results further showed that tactile feedback was perceived as more useful for multi-surface swabbing and defect detection tasks than for obstacle avoidance ($F(2,40)=3.33$, $p=.046$), which is intuitive as perception of the tactile feedback cues does not increase as linearly as visual feedback.

Overall, the results indicate that the primary advantage of tactile feedback over visual feedback lies in enhancing telepresence and reducing response latency rather than reducing cognitive workload, with the greatest benefit observed in 
contact-rich, multi-surface manipulation tasks. By offering a low-cost, non-wearable touchscreen interface that delivers force feedback through a natural tactile modality, this work contributes to the design of more intuitive and effective robot teleoperation interfaces.

This study advances the understanding of feedback modality effects in teleoperation and provides a practical method for improving operator situational awareness through enhanced telepresence. It introduces a teleoperation interface enabling intuitive control for surface interaction tasks, with direct applicability to domains such as nuclear decommissioning, surgical robotics or space robotics, where precise surface operations are required.

Based on the observations, key advantages of integrating electrovibration as a tactile feedback method for teleoperation are as follows:

\begin{itemize}
    \item The signal delay between generation and perception is extremely low, as feedback propagates at electrical speed with no mechanically moving parts, unlike vibration motors or piezoelectric actuators, addressing one of the primary barriers to tactile telepresence in robotics.
    \item The feedback modality directly matches the modality of tactile perception on the robot side, eliminating the need for sensory substitution and allowing operators to perceive remote surface interactions naturally.
    \item The method does not suffer from the stability issues reported in the literature for kinesthetic feedback during stiff environment contacts, making it more robust in practical teleoperation scenarios.
    \item The approach effectively renders remote environments with potential extensions to medical telerobotics, assistive technologies for people with motor impairments, and remote operations requiring enhanced situational awareness.
\end{itemize}

The key disadvantages and current limitations of the approach are:

\begin{itemize}
    \item Electrovibration requires relative motion between the fingertip and the screen for a sensation to be perceived, which prevents feedback during static contact.
    \item The reflected friction force is perceivable but does not map one-to-one to the environment contact force, which constrains force fidelity.
    \item The current implementation requires an antistatic wristband attached to the operator’s hand, which slightly restricts hand movement. Differential voltage-output hardware was used in this study to enhance safety in the event of a short circuit, however, with absolute voltage-output hardware, the wristband is not necessary.
\end{itemize}

A notable observation from the study was that the level of tactile perception varied across participants. While most participants perceived the feedback immediately, others required time for sufficient charge to build up at the fingertip. Three participants initially reported no perception of the feedback,however, upon being electrically grounded by touching the metal casing of a nearby computer, all three began perceiving it. This suggests that body grounding plays a significant role in electrovibration perception, and future interface designs should integrate a grounding mechanism to ensure consistent perception across all users.

Limitations of this study include the relatively small sample size of 21 participants, which may affect the generalisability of the results, and the use of a simulated robot in a laboratory setting with non-expert operators, which limits direct applicability to industrial practice. A real robotic setup would use the same operator interface and tactile feedback hardware, but the robot-environment side may introduce additional network delay and sensor noise, which could affect the timing and reliability of the tactile cues. 

Future work will focus on testing the interface with real robots and domain experts to assess robustness and usability in operational settings, including the effect of communication delay and sensor noise on tactile feedback quality, as well as integrating grounding into the hardware design and exploring the combination of tactile and visual feedback modalities.

\section{Conclusion}
\label{sec:Conclusions}

This study designed, implemented, and evaluated a touchscreen-based teleoperation interface that uses electrovibration to provide tactile feedback without notable delays. Characterisation experiments confirmed that the interface can convey force outputs at sufficiently low latency, with all participants able to perceive the tactile feedback, and faster response times under the tactile feedback compared to visual feedback.

A comparative study with 21 participants across two feedback modality conditions demonstrated that tactile feedback significantly enhances the sense of telepresence and reduces operator response latency, while imposing comparable cognitive load to visual feedback. Task ranking results further highlighted that the benefits of tactile feedback are most pronounced in contact-rich surface interaction tasks, such as multi-surface swabbing and defect detection.

The proposed interface provides a practical and low-cost solution for enhancing precision and situational awareness in surface interaction tasks. By enabling operators to perceive remote surface interactions naturally through the fingertip, it helps reduce the risk of errors in hazardous environments. This research demonstrates that electrovibration-based tactile telepresence can enable safer and more intuitive robotic manipulation in critical industrial applications.

\section*{Acknowledgement}

This work was supported by the European Commission’s Marie Skłodowska-Curie Actions (MSCA) Project RAICAM (GA 101072634), and UK Research and Innovation (UKRI) grant number EP/X025977/1.

\balance

\bibliographystyle{IEEEtran}  
\bibliography{IEEEabrv}

\vspace{12pt}

\end{document}